\documentclass[letterpaper, 10 pt, journal, twoside]{IEEEtran}  

\usepackage[T1]{fontenc}
\usepackage{booktabs}
\usepackage{tabularx}
\usepackage{tikz}

\IEEEoverridecommandlockouts                              

\usepackage[
backend=biber,
style=ieee,
dashed=false,
citestyle=numeric-comp
]{biblatex}

\DefineBibliographyStrings{english}{
  andothers = {et\addabbrvspace al\adddot}
}

\usepackage[acronym]{glossaries}
\usepackage{caption}

\usepackage{siunitx}
\newacronym{MPC}{MPC}{Model Predictive Control}
\newacronym{MHE}{MHE}{Moving Horizon Estimation}
\newacronym{RBFN}{RBFN}{Radial Basis Function Networks}
\newacronym{OCP}{OCP}{Optimal Control Problem}

\usepackage{graphicx}
\usepackage{subcaption}
\usepackage{microtype}
\usepackage{amsmath} 
\usepackage{amssymb}  
\usepackage{amsmath,amssymb,amsfonts}
\usepackage{tensor}
\usepackage{tablefootnote}
\usepackage{lipsum}

\usepackage{hyperref}
\hypersetup{
    colorlinks,
    linkcolor={black},
    citecolor={black},
    urlcolor={black}
}

\DeclareMathOperator{\joint}{\mathbf{q}}

\DeclareMathOperator{\taskerror}{\mathbf{e}}
\DeclareMathOperator{\state}{\mathbf{x}}
\DeclareMathOperator{\control}{\mathbf{u}}
\DeclareMathOperator{\desired}{\text{ref}}
\DeclareMathOperator{\surfaceweights}{\boldsymbol{\theta}}
\DeclareMathOperator{\surface}{s}

\DeclareMathOperator{\slack}{\boldsymbol{\varepsilon}}

\title{
From Instantaneous to Predictive Control: 
A More Intuitive and Tunable MPC Formulation for Robot Manipulators
}

\author{Johan Ubbink$^{1}$, Ruan Viljoen$^{1}$, Erwin Aertbeli\"en$^{1}$, Wilm Decr\'e$^{1}$, and Joris De Schutter$^{1}$
\thanks{Manuscript received: July, 17, 2024; Revised October, 11, 2024; Accepted November, 19, 2024.}
\thanks{This paper was recommended for publication by Editor Jaydev P. Desai upon evaluation of the Associate Editor and Reviewers' comments.}
\thanks{This work was supported by project G0D1119N of the Research Foundation - Flanders (FWO - Flanders).} 
\thanks{$^{1}$All authors are with the Department of
Mechanical Engineering, KU Leuven, and Flanders Make@KU Leuven, Leuven, Belgium.
        {\tt\footnotesize johan.ubbink@kuleuven.be}}%
\thanks{Digital Object Identifier (DOI): see top of this page.}
}

\begin{document}

\markboth{IEEE Robotics and Automation Letters. Preprint Version. Accepted November, 2024}
{Ubbink \MakeLowercase{\textit{et al.}}: A More Intuitive and Tunable MPC Formulation for Robot Manipulators}  

\maketitle

\begin{abstract}

Model predictive control (MPC) has become increasingly popular for the control of robot manipulators due to its improved performance compared to instantaneous control approaches.
However, tuning these controllers remains a considerable hurdle. 
To address this hurdle, we propose a practical MPC formulation which retains the more interpretable tuning parameters of the instantaneous control approach while enhancing the performance through a prediction horizon.
The formulation is motivated at hand of a simple example, highlighting the practical tuning challenges associated with typical MPC approaches and showing how the proposed formulation alleviates these challenges.
Furthermore, the formulation is validated on a surface-following task, illustrating its applicability to industrially relevant scenarios.
Although the research is presented in the context of robot manipulator control, we anticipate that the formulation is more broadly applicable.
\end{abstract}

\section{Introduction} \label{sec:introduction}
\IEEEPARstart{W}{hen} deploying robot manipulators in unstructured environments, it is necessary to account for disturbances.
Traditionally, this has been achieved by implementing instantaneous feedback controllers in the task space that continuously correct for disturbances \cite{samson, etasl}, referred to in this paper as the instantaneous control approach.
However, as this approach is purely instantaneous, it fails to anticipate future constraints such as workspace and actuator limits, resulting in suboptimal performance.
To address this limitation, there has been an increased interest in using model predictive control (MPC), which can better anticipate constraints by incorporating a prediction horizon \cite{Cefalo:2018, gold:2023, jordana2024}.

However, tuning these MPC controllers is a known challenge~\cite{Grandia:2019}, which is particularly problematic for high-mix, low-volume applications where the robot is frequently reprogrammed for new tasks. 
If the controller is difficult to tune, it significantly hinders the economic viability of using MPC. 
Therefore, these applications can benefit from an MPC formulation that is more intuitive and easier to tune.

To simplify MPC controller tuning, recent work has explored learning-based methods \cite{Mehndiratta:2018, Edwards:2021}.
Although promising, several open challenges remain. 
Firstly, many of these approaches rely on a highly accurate simulation environment, which is often unavailable. 
Secondly, these approaches optimise a higher-level objective or reward function with additional hyperparameters, which must also be tuned. 
Finally, when there are many tuning parameters, these approaches tend to struggle as the search space becomes prohibitively large. 
Thus, starting with an MPC formulation that is easier to tune can reduce the search space and is therefore also beneficial for these approaches.

As explained by \textcite{Grandia:2019}, much of the tuning effort can be avoided by limiting the bandwidth of the MPC controller, which helps prevent the excitation of unmodeled dynamics.
For this reason, they present an MPC formulation that penalises the high-frequency components of control inputs within the objective function. 
In this way, they avoid the need to accurately model every detail of the robot and its interaction with the environment.

Limiting the controller's bandwidth to reduce the tuning effort is also a common practice when using the previously mentioned instantaneous control approach \cite{samson, etasl}, where the bandwidth is limited by directly specifying the desired dynamics for each task error (e.g. in terms of a time constant).
Additionally, based on these specified time constants, application developers can anticipate the resulting behaviour of the controller, which ensures a safer tuning process.
In contrast, with typical MPC formulations, it is difficult to anticipate the closed-loop behaviour of the controller directly from the tuning parameters, as will be demonstrated in Section~\ref{sec:example}.

Another consideration is that for many applications there may already be an existing instantaneous controller. 
What if the application developer wants to improve the performance by using an MPC controller? 
Should they completely discard the existing controller and start from scratch, or is there a way to more easily transition from the instantaneous controller to an MPC controller?

In light of this, \textcite{Cairano:2010} presented an approach where an MPC controller is tuned to behave like a given linear state-feedback controller. 
This approach enables the MPC controller to inherit the stability, robustness, and frequency properties of the linear controller around the equilibrium where constraints are inactive, while leveraging the improved performance of MPC during transient responses when the constraints are active. 
However, as their approach is limited to approximating linear state-feedback controllers, it is not directly applicable to the control of robot manipulators, which typically require incorporating the nonlinear forward kinematics of the robot.

In contrast, \textcite{Bednarczyk:2020} presented a similar approach, but targeted at robot manipulators.
However, their focus was limited to combining impedance control and MPC, while we aim to address a wider class of problems.
Moreover, their formulation solves a linearized MPC problem at each time step, while we propose solving the full nonlinear MPC problem, which is known to offer better performance.

Concretely, our approach focuses on the practical tuning challenges faced when applying typical MPC formulations to robot control applications.
Inspired by the instantaneous control approach, we propose an MPC formulation where it is possible to directly specify a desired time constant for each task error.
This formulation results in tuning parameters which are more intuitive and can be selected to limit the closed-loop bandwidth.
Furthermore, the formulation is validated on a surface-following
task, illustrating its applicability to industrially relevant scenarios.
More specifically, we:
\begin{itemize}
\item illustrate the challenges related to MPC controller tuning using a simple 2D example (Section~\ref{sec:example});
\item provide a short introduction to instantaneous robot control and present an MPC formulation inspired by this approach (Section~\ref{sec:instantaneous} and ~\ref{sec:decay_mpc});
\item demonstrate the approach on a surface-following task, showing how the intuition behind the instantaneous control approach can be retained while improving performance through MPC (Section~\ref{sec:surface_following}).
\end{itemize}

While this research is presented in the context of robot manipulator control, the formulation may be relevant for MPC practitioners in other domains, as the challenges related to tuning are not limited to robotics alone.

\section{Motivating Example} \label{sec:example}
\begin{figure}
    \centering
    \includegraphics[width = 0.39\textwidth]{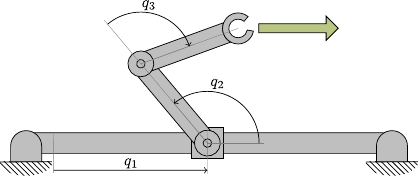}
    \caption{Scenario where a 3-DOF robot manipulator is controlled to follow a Cartesian trajectory.}
    \label{fig:arm}
\end{figure}

Fig.~\ref{fig:arm} shows a scenario where the end-effector of a 3-DOF planar robot should follow a Cartesian trajectory.
The source code and animations related to this example are available 
at {\small \url{https://github.com/johanubbink/tuning_mpc}}.

For this example, the robot is assumed to be an ideal acceleration-controlled system with control input $\control = [\ddot{\joint}_\text{cmd}]$, i.e. the real acceleration is assumed to be equal to the commanded one.
The system's state $\state = [\joint^\intercal , \dot{\joint}^\intercal, t ]^\intercal$ consist of the joint positions $\joint = [q_0, q_1, q_2]^\intercal$, joint velocities, and time. The latter is included to simplify the notation throughout the rest of the paper. 
The continuous-time system dynamics are written as:
\begin{equation} \label{eq:acceleration_system}
    \dot{\state} = \mathbf{f}_\text{c}(\state, \control) = 
    \begin{bmatrix}
        0 & 1 & 0 \\
        0 & 0 & 0 \\
        0 & 0 & 0 \\
    \end{bmatrix}
    \begin{bmatrix}
        \joint \\ \dot{\joint} \\ t
    \end{bmatrix}
    +
    \begin{bmatrix}
        0 & 0 \\
        1 & 0 \\
        0 & 1 
    \end{bmatrix}
    \begin{bmatrix}
        \ddot{\joint}_\text{cmd}  \\ 1
    \end{bmatrix}.
\end{equation}

The goal is to regulate the task error $\mathbf{e}(\state)$ between the robot's end-effector and a desired sinusoidal trajectory:
\begin{align} \label{eq:position_task}
    \underbrace{
    \begin{bmatrix}
        e_1 \\ e_2
    \end{bmatrix}
    }_{\mathbf{e}(\state)}
     &= 
     \underbrace{
    \begin{bmatrix}
    q_1 + \cos{(q_2)} + \cos{(q_2 + q_3)}\\
     \sin{(q_2)} + \sin{(q_2 + q_3)}
    \end{bmatrix}}_{\text{end-effector position}}
    -
    \underbrace{
    \begin{bmatrix}
    \cos (t) \\
    1.5 
    \end{bmatrix}
    }_{\substack{\text{desired} \\ \text{trajectory}}}.
\end{align}
To achieve this goal, an MPC controller is designed by specifying the following optimal control problem (OCP) and solving it at every control interval:
\begin{subequations} \label{eq:mpc}
        \begin{align} 
                \min_{\control, \state} \, \, & 
                \sum_{k=0}^{N-1} l(\state_k,\control_k) \,\, + V_N(\state_N) \label{eq:ocp_obj}
                \\[2ex]
                \text{s.t.} \quad
                & \state_{k+1} = \mathbf{f}_\text{d}(\state_k, \control_k). \label{eq:ocp_sys_dynamics}
        \end{align}
\end{subequations}
This OCP minimises an objective function in \eqref{eq:ocp_obj} over a horizon length of $N$, subject to the discrete-time system dynamics in \eqref{eq:ocp_sys_dynamics}. The discrete-time system dynamics $\mathbf{f}_\text{d}$ is obtained by discretizing the continuous-time dynamics in \eqref{eq:acceleration_system}. For simplicity, this example uses forward-Euler discretization, however, it is also possible to use more advanced approaches~\cite{Rawlings2017}. 
The objective function consists of a stage cost $l$ and a terminal cost $V_N$, which describes the desired behaviour of the controller.
In theory, a stable response can be guaranteed by selecting an appropriate terminal cost~\cite{Rawlings2017}. 
However, in practice, this is challenging and is often omitted. 
Therefore, in this example, we also omit the terminal cost.

Practitioners often start with a specific objective function and tune the weights and structure through trial and error, which can be tedious and time-consuming\footnote{Jokingly referred to as ``graduate student descent''}. 
In the remainder of this section, we will illustrate this challenge using two possible objective functions, A and B.

\subsection{Objective function A} \label{sec:tracking_controller}
Although there is no ``standard'' MPC formulation, it is common to specify an objective function inspired by the linear quadratic regulator (LQR) with a stage cost:
\begin{equation} \label{eq:mpc_objective}
        l^\text{A} (\state_k,\control_k)
        =
        \mathbf{e}(\state_k)^\intercal \mathbf{Q} \mathbf{e}(\state_k) + \control_k^\intercal \mathbf{R} \control_k.
\end{equation}
This objective function trades off the task error $\mathbf{e}$ with the control input $\control$, weighted with $\mathbf{Q}$ and $\mathbf{R}$ respectively. 

\begin{figure}
    \centering
    \begin{subfigure}[b]{0.45\textwidth}
        \includegraphics[width=\textwidth]{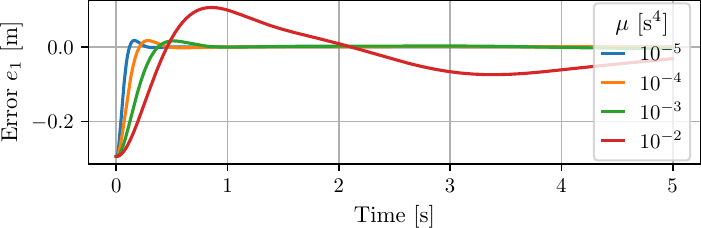}
        \caption{Perfect simulation model.}
        \label{fig:objective_1}
    \end{subfigure}
    \hfill
    \begin{subfigure}[b]{0.45\textwidth}
        \includegraphics[width=\textwidth]{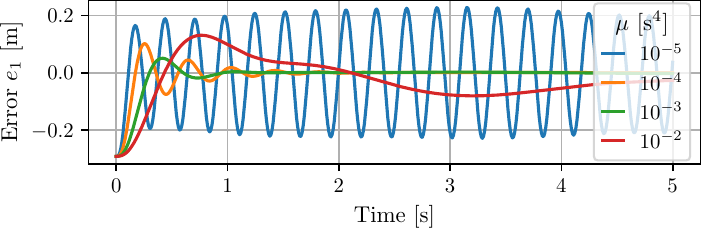}
        \caption{Simulation model with actuator dynamics.}
        \label{fig:objective_1_dynamics}
    \end{subfigure}
    \caption{Tracking error $e_1$ for different $\mu$ values using objective function A.}
    \label{fig:combined_objective_1}
\end{figure}

How to choose appropriate values for the weights is not immediately obvious. 
To illustrate this point, we selected $\mathbf{Q} = \text{diag}(\SI{1}{\per \meter \squared}, \SI{1}{\per \meter \squared})$ and $\mathbf{R} = \mu \mathbf{W}_r$, with $\mathbf{W}_r = \text{diag}(\SI{1}{\per \meter \squared}, \SI{1}{\per \radian \squared}, \SI{1}{\per \radian \squared})$  where $\mu$ adjusts the level of regularisation. 
The closed-loop system was simulated for different values of $\mu$, with $N=40$ and $\Delta t = \SI{0.01}{\second}$.
Fig.~\ref{fig:objective_1} shows the tracking error $e_1$, where it can be observed that for small values of $\mu$, the error quickly decays, perfectly tracking the desired trajectory defined in \eqref{eq:position_task}.
When $\mu$ is increased, the error decays more slowly and the tracking performance degrades.

However, this simulation used the same model as the MPC controller, without any mismatch.
For a more realistic scenario, the simulation environment is modified to include a first-order model representing internal actuator dynamics:
\begin{equation} \label{eq:internal_dynamics}
    \ddot{\joint}_{k+1} = \ddot{\joint}_{k} - \Delta t  \, \alpha_\text{internal} ( \ddot{\joint}_{k} - \ddot{\joint}_{\text{cmd},k}).
\end{equation}

Fig.~\ref{fig:objective_1_dynamics} shows the results for the same experiments, now simulated with the model mismatch where $\alpha_\text{internal} = \SI{15}{\per \second}$. 
The introduced actuator dynamics cause the very fast response ($\mu = 10^{-5}~\si{\second^{4}}$) to become unstable, whereas the slower responses are less affected by the mismatch.

To account for this mismatch, there are multiple options. 
The first option is to tediously perform system identification to increase the accuracy of the MPC model. 
However, this can be very time-consuming as the exact cause is often unknown, stemming from internal robot dynamics, a sensor, interaction with the environment, or a combination of these variables.
A simpler, though more conservative, approach is to decrease the closed-loop bandwidth of the MPC controller (i.e. make the controller less aggressive) to avoid exciting the unmodeled system dynamics. 
This approach is typically used in more classical control approaches and is common practice in industrial robot controllers.

\subsection{Objective function B}
The closed-loop bandwidth of the MPC controller can be decreased by reducing the speed of the response. 
This can be achieved by penalising the velocity of the system in the objective function, which can be interpreted as adding a damping term. 
Penalising the velocities can be done in either the joint space, end-effector space, or task space.
However, penalising the joint velocity introduces unpredictable behaviour in the task space, and penalising the end-effector velocity leads to tracking errors.
Therefore, we choose to penalise the task velocity $\dot{\taskerror}$ within the following stage cost:
\begin{equation}\label{eq:mpc_objective_2}
        l^\text{B} (\state_k,\control_k)
        =
        \begin{bmatrix} 
            {\taskerror}(\state_k) \\ \dot{\taskerror}(\state_k)
        \end{bmatrix}^\intercal
        \mathbf{Q}^\text{B}
        \begin{bmatrix} 
            {\taskerror}(\state_k) \\ \dot{\taskerror}(\state_k)
        \end{bmatrix}
        + \control_k^\intercal \mathbf{R} \control_k,
\end{equation} 
where $\dot{\taskerror}$ is computed by applying the chain rule and substituting in the continuous-time system dynamics:
\begin{equation}
    \dot{\taskerror}(\state)  = \frac{\partial \taskerror}{\partial \state} \mathbf{f}_\text{c}(\state, \control).
\end{equation}

\begin{figure}
    \centering
    \includegraphics[width = 0.45\textwidth]{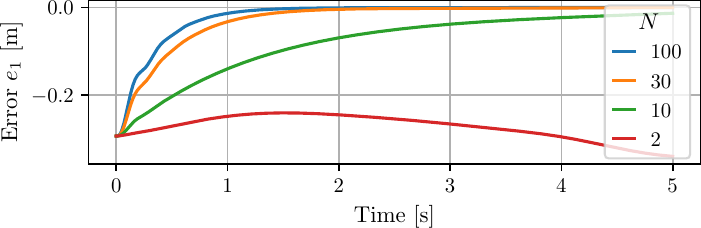}
    \caption{Tracking error $e_1$ for different horizon lengths $N$ using objective function B.}
    \label{fig:objective_2}
\end{figure}

As before, it is not immediately clear how the choice of weights affects the closed-loop response of the system.
For this example we selected diagonal matrices $\mathbf{Q}^\text{B} = \text{diag}(\mathbf{I}, \lambda \mathbf{I})$ and $\mathbf{R} = \mu \mathbf{W}_r$, where $\lambda$ adjusts the regularisation of the task velocity. 
The controller was hand-tuned for values of $\mu = 10^{-4}~\si{\second^{4}}$, $\lambda = 10^{-1}~\si{\second^{2}}$, and $N=30$ to achieve a stable response with small tracking errors as shown in Fig.~\ref{fig:objective_2}. 

However, the figure also shows the resulting closed-loop response for different horizon lengths, which drastically changes as the horizon length varies.
For example, the closed-loop response for $N=10$ is much slower than for $N=30$, and for $N=2$, the end-effector completely fails to reach the target. 
If the horizon length is modified for any reason, such as being decreased to reduce computation time or increased to better anticipate constraints, it might be necessary to repeat a tedious tuning procedure.

The example presented here is relatively trivial, with no input or state constraints, only three degrees of freedom, and no sensor noise. 
If tuning is already challenging for this simple example, it is expected to be much more difficult for real-world applications.

\section{Instantaneous Control Approach} \label{sec:instantaneous}

As demonstrated with the motivating example, anticipating the resulting closed-loop dynamics of an MPC controller is difficult when using the typical tuning parameters. 
In contrast, the instantaneous control approach allows for the direct specification of the desired closed-loop dynamics, such as the following first-order response:
\begin{equation} \label{eq:vel_resolved}
    \frac{\text{d}}{\text{d} t}{\taskerror}(\state) = -\mathbf{K}_e \taskerror(\state).
\end{equation}
The gain $\mathbf{K}_e$ is typically chosen as a diagonal matrix, i.e. $\mathbf{K}_e = \text{diag}(\alpha_1, \alpha_2, \dots) $. 
The advantage of this approach is that the tuning parameters $\alpha_i$ are directly interpretable: (a) the time constant $\alpha_i^{-1}$ provides insight into how quickly the corresponding task error $e_i$ decays, and (b) it can be selected to limit the closed-loop bandwidth of the controller.
In practice, the value of $\alpha_i$ is chosen as large as possible without exciting the unmodeled dynamics of the system.
It is advisable to start with a conservative gain value and gradually increase it. 
If instabilities emerge (e.g. the robot oscillates), the gain should be reduced (by about half), similar to the approach proposed by \textcite{Ziegler1942OptimumSF} in 1942.

Assuming a velocity-controlled robot as in \cite{etasl}, where the state is given by $\state = [\joint^\intercal, t]^\intercal$, the control input $\control = \dot{\joint}$ can be found by expanding the left-hand side of \eqref{eq:vel_resolved} as follows:
\begin{align} \label{eq:ten}
   \underbrace{
   \frac{\partial \taskerror(\joint, t)}{ \partial \joint} \dot{\joint} 
    + \frac{\partial \taskerror(\joint, t)}{ \partial t}
    }_{\dot{\taskerror}(\state, \control)}
    = -\mathbf{K}_e \taskerror(\state).
\end{align}
A similar approach can be used for an acceleration-controlled robot by specifying a second-order response for tasks of relative degree two (e.g. position-level tasks) \cite{bouyarmane_2018}.

As in \cite{etasl}, the control input can be found by embedding \eqref{eq:ten} within an optimisation problem:
\begin{subequations}\label{eq:vel_instant}
    \begin{align} 
        \min_{{\control, \slack}} & \quad
        \slack^\intercal \mathbf{W}_s \slack
        +
        \mu \control^\intercal \mathbf{W}_r \control \label{eq:instant_objective}
        \\
        \text{s.t.}
        & \quad \dot{\taskerror}(\state, \control) + \mathbf{K}_e \taskerror(\state) = \slack_e \label{eq:equality}, \\
        & \quad \dot{\mathbf{h}}(\state, \control) + \mathbf{K}_h \mathbf{h}(\state) \geq \slack_h \label{eq:inequality} .
    \end{align}
\end{subequations}
By adding slack variables $\slack = [\slack_e^\intercal, \slack_h^\intercal]^\intercal$ and weighting them in the objective with $\mathbf{W}_s$, it is possible to address conflicting \textit{soft} constraints.
For \textit{hard} constraints, the corresponding slack variable can be replaced with zero. 
The control input is also regularised in the objective function, but is multiplied by a very small weight $\mu$. 
As a result, $\mathbf{W}_r$ only significantly influences the control when there are more degrees of freedom than tasks, leading to the solution with the smallest control effort, or when the robot is near a singular configuration, where large control inputs are penalised.

The formulation also allows for inequality constraints,
\begin{equation} \label{eq:simple_inequality}
    \mathbf{h}(\state) \geq 0,
\end{equation}
by using \eqref{eq:inequality}, where the inequalities are treated similar to the task errors.
By enforcing the inequalities in this way, the controller avoids exciting the previously mentioned unmodeled dynamics when the inequality constraints are suddenly activated.
Similar to $\mathbf{K}_e$, $\mathbf{K}_h$ is selected as a diagonal matrix with entries $\alpha_i$, which can be chosen such that the robot does not approach the bound of a constraint faster than the bandwidth of the system.
This inequality can also be interpreted as a Control Barrier Function (CBF)~\cite{Ames:2014}, an increasingly popular approach in nonlinear control for ensuring a system's safety.

\section{Proposed MPC formulation} \label{sec:decay_mpc}
We propose an MPC formulation that retains the more intuitive tuning parameters of the instantaneous control approach, while improving the performance by extending it with a prediction horizon.

\subsection{Objective function}
Inspired by the instantaneous control approach in \eqref{eq:vel_instant}, we propose the following stage cost:
\begin{equation} \label{eq:pdmpc_objective}
        l^\text{C}(\state_k,\control_k) = {\slack}(\state_k, \control_k)^\intercal \mathbf{W}_s {\slack}(\state_k, \control_k)
        + \mu \control_k^\intercal \mathbf{W}_r \control_k,
\end{equation}
where $\slack(\state,\control)$ is defined as
\begin{align} \label{eq:decay_slack}
    \slack(\state,\control) = \dot{\taskerror}(\state, \control) + \mathbf{K}_e \taskerror(\state),
\end{align}
and represents the deviation from a first-order response. 
Here, $\taskerror$ is a function of only $\state$, but $\dot{\taskerror}$ can depend on $\control$, depending on the relative degree of $\taskerror$.

Interestingly, by substituting \eqref{eq:decay_slack} into \eqref{eq:pdmpc_objective} it is possible to show that this objective is equivalent to \eqref{eq:mpc_objective_2} when selecting:
\begin{equation}
    \mathbf{Q}^\text{B} = 
    \begin{bmatrix}
    \mathbf{K}_e^\intercal \mathbf{W}_s \mathbf{K}_e &  \mathbf{K}_e^\intercal \mathbf{W}_s \\
     \mathbf{W}_s \mathbf{K}_e & \mathbf{W}_s
    \end{bmatrix}, \quad
    \mathbf{R} = \mu \mathbf{W}_r.
\end{equation}
Therefore, this formulation can be viewed as a way of choosing $\mathbf{Q}^\text{B}$ and $\mathbf{R}$ with a specific structure that keeps the intuition of the instantaneous control approach. That is, the ability to directly specify the time constant $\alpha^{-1}$ for each task error, which can be selected to limit the closed-loop bandwidth of the MPC controller.

\subsection{Inequality constraints} \label{sec:inequalities_mpc}

To incorporate inequality constraints into the MPC formulation, two approaches have been considered.
The first and most common approach is to directly enforce the inequality constraints within the OCP (i.e. $\mathbf{h}(\state) \geq 0$).
An alternative approach is to use the CBF-type constraint in \eqref{eq:inequality}. 
This alternative approach has been investigated and compared to the typical approach in \cite{Son:2019,allamaa:2024}, showing that enforcing the inequalities in this way improved safety, especially in the presence of unmodeled dynamics and when using shorter prediction horizons. 
Therefore, we choose the second approach. 
Similar to the instantaneous approach, these constraints can either be hard or soft, depending on whether the corresponding slack variables are included or excluded.

To better understand the relationship between the proposed constraint formulation and enforcing the constraints directly in the OCP, we can consider a single constraint with no slack variables. It is possible to rewrite the constraint as:
\begin{equation}
    \frac{1}{\alpha_i} \dot{h}_i(\state,\control) + h_i(\state) \geq 0.
\end{equation}
If the tuning parameter $\alpha_i$ is chosen to be infinitely large, then the original constraint is recovered.
Therefore, directly enforcing the constraint in the OCP implicitly assumes that the system can react infinitely fast, which is unrealistic for a real setup with unmodeled dynamics.

\subsection{Revisiting the motivating example}
\begin{figure}
    \centering
    \begin{subfigure}[b]{0.45\textwidth}
        \includegraphics[width=\textwidth]{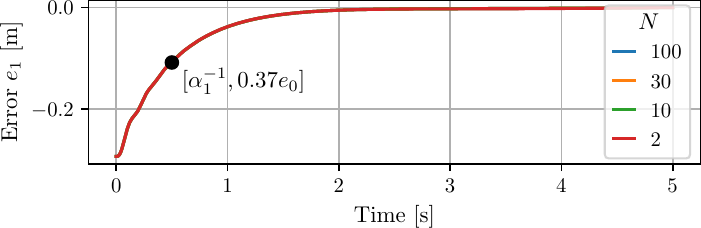}
        \caption{For $\mu = 10^{-5}~\si{\second^{4}}$ with different horizon lengths $N$.}
        \label{fig:objective_3}
    \end{subfigure}
    \hfill
    \begin{subfigure}[b]{0.45\textwidth}
        \vspace{5pt}
        \includegraphics[width=\textwidth]{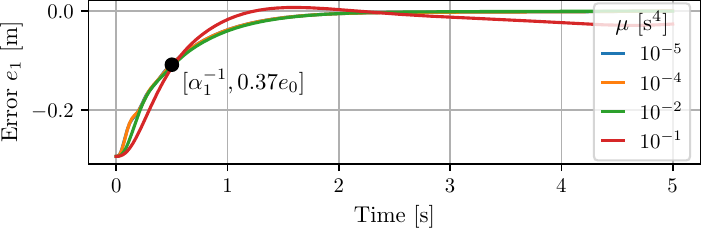}
        \caption{For $N=40$ with different regularisation levels $\mu$.}
        \label{fig:objective_3_mu_sweep}
    \end{subfigure}
    \caption{Tracking error $e_1$ with the proposed objective function.}
    \label{fig:combined_objective_3}
\end{figure}

To illustrate the advantage of the proposed objective function, it is applied to the motivating example from Section~\ref{sec:example}, and simulated with the model mismatch introduced in \eqref{eq:internal_dynamics}.
Fig.~\ref{fig:objective_3} shows the tracking error $e_1$ for different horizon lengths with $\mathbf{K}_e = \text{diag}(\SI{2}{\per \second},\SI{2}{\per \second})$ , $\mathbf{W}_s = \text{diag}(\SI{1}{\second \squared \per \meter \squared}, \SI{1}{\second \squared \per \meter \squared})$, $\mathbf{W}_r = \text{diag}(\SI{1}{\per \meter \squared}, \SI{1}{\per \radian \squared}, \SI{1}{\per \radian \squared})$, and $\mu = 10^{-5}~\si{\second^{4}}$.

From the figure, it is evident that, regardless of the horizon length, all simulations closely followed the desired exponential response with a time constant of $\alpha_1^{-1} = \SI{0.5}{\second}$. 
The slight deviation from a perfect response is due to the mismatch between the MPC model and the simulation model. 
However, this effect is negligible, as the bandwidth of the MPC controller was selected to be significantly lower than that of the unmodeled internal actuator dynamics.

The behaviour is appealing for two reasons: Firstly, unlike in Fig.~\ref{fig:objective_2}, the transient response does not change when the horizon length changes.
This property makes the tuning easier, as it is less dependent on the selected horizon length.
The second appealing property is that the interpretation of the tuning parameter $\alpha_1 = \SI{2}{\per \second}$ is clear. 
Within the time frame of $\alpha_1^{-1} = \SI{0.5}{\second}$, the error reduces to $37\%$ of the initial error $e_0$.
This property provides the application developer with an understanding of how the MPC controller should behave even before running it, making it safer to tune in the real world.

With the proposed formulation, it is still necessary to select an appropriate value for $\mu$.
However, unlike with objective function $l^{A}$ where $\mu$ is used to adjust the aggressiveness of the controller, here it is added mainly for numerical reasons to ensure that the optimisation problem remains well conditioned.
Furthermore, we observed that the resulting behaviour of the controller is not that sensitive to the choice of $\mu$.
For example, Fig~\ref{fig:objective_3_mu_sweep} shows the tracking error simulated for different values of $\mu$, assuming a horizon of $N=40$. 
As long as $\mu$ is ``small enough" ($\mu \leq 10^{-2}~\si{\second^{4}}$), the resulting closed-loop dynamics are not greatly affected.

\begin{figure}
    \centering
    \includegraphics[width = 0.45\textwidth]{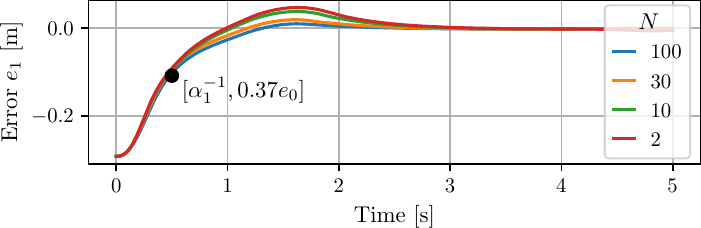}
    \caption{Tracking error $e_1$ with the proposed objective function with constraints.}
    \label{fig:objective_3_with_constraints}
\end{figure}

To demonstrate the constraint formulation and benefit of a prediction horizon, the example in Section~\ref{sec:example} is extended with the following state constraints:
\begin{gather}
    \joint^{-} \leq \joint \leq \joint^{+}, \label{eq:pos_limit} \\
    \dot{\joint}^{-} \leq \dot{\joint} \leq \dot{\joint}^{+}, \label{eq:vel_limit}
\end{gather}
where $\square^{-}$ and $\square^{+}$ denote the lower and upper limits.
These state constraints, which can be written in the form of \eqref{eq:simple_inequality}, are imposed as hard constraints in the OCP by using the barrier-type formulation from \eqref{eq:inequality} with $\alpha_i = \SI{2}{\per \second}$ for all constraints. Input constraints are also enforced:
\begin{equation}
    \ddot{\joint}^{-} \leq \ddot{\joint}_\text{cmd} \leq \ddot{\joint}^{+}. \label{eq:accel_limit}
\end{equation}
However, as this constraint consists of the control inputs, it is directly enforced, without specifying any dynamics.

Fig.~\ref{fig:objective_3_with_constraints} shows the tracking error for different horizon lengths.
Compared to Fig.~\ref{fig:objective_3}, the error no longer follows a perfect exponential response. 
This deviation is due to the constraints imposed on the OCP, which make it impossible to follow the desired response. 
However, the advantage of including the prediction horizon becomes apparent. 
By increasing the horizon length, the controller can better anticipate the constraints and react earlier, reducing tracking errors. 

\subsection{Specifying higher-order dynamics}
As mentioned in Section~\ref{sec:instantaneous}, when using an instantaneous control approach with an acceleration-controlled robot, it is necessary to specify second-order dynamics. 
In contrast, when using the proposed MPC formulation, it is possible to specify first-order dynamics as in \eqref{eq:decay_slack}, as long as a prediction horizon of $N \geq 2$ is used.
It is also possible to specify second-order dynamics such that for a horizon of $N=1$ the controller is exactly the same as an instantaneous controller. 
However, we found that imposing these higher-order dynamics along the prediction horizon is more computationally expensive, with no additional benefit in terms of tuning.
Therefore, we do not address it further in this paper.

\subsection{Comparison with CLF-MPC} \label{sec:clf_mpc}
In the field of nonlinear control theory, a control approach known as the Pointwise Min-Norm Controller \cite{Freeman:1995} has been presented which closely resembles the instantaneous control approach in \eqref{eq:vel_instant}.
Based on Control Lyapunov Functions (CLFs), a scalar positive definite function $V$ is defined and used as a measure of the ``energy'' of the system. 
If $V$ is a valid CLF, stability can be achieved by using the following control law:
\begin{subequations} \label{eq:clf_qp}
\begin{align} 
    \control = \arg \min_{\control} & \quad
    \control^\intercal \control
    \\
    \text{s.t.}
    & \quad \dot{V}(\state, \control) \leq - \alpha V(\state), \label{eq:clf} 
\end{align}
\end{subequations}
given that the optimisation problem is feasible for all possible system states.
This control law selects the smallest control input that satisfies the inequality in \eqref{eq:clf}. 

Similarly to how our approach extends the instantaneous controller in \eqref{eq:vel_instant} to a prediction horizon, this Pointwise Min-Norm Controller has been extended to a prediction horizon \cite{Primbs:2000, GrandiaCLF:2020}. 
However, unlike these approaches, our formulation aligns more closely with the instantaneous control approach presented in Section~\ref{sec:instantaneous} where the desired dynamics are directly specified for each task error.
This choice eliminates the need to find a valid CLF and reduces the number of inequality constraints, as the errors are directly penalised in the objective. 
Nonetheless, we observed similar advantages to those found in research combining CLFs with MPC: easier tuning, particularly for short horizons where the controller behaves like the instantaneous equivalent, and improved performance with longer horizons.

\section{Surface Following Example} \label{sec:surface_following}
To demonstrate the proposed MPC formulation, it was applied to the contactless surface-following task previously investigated in \cite{ubbink2024}. 
The previous study highlighted MPC's advantages for surface following, particularly in handling constraints, though it was only tested in simulation. 
While the typical LQR-inspired objective function in \eqref{eq:mpc_objective} performed well in simulation, real-world implementation and tuning were challenging. This challenge inspired our research, leading to the exploration of more intuitive MPC formulations.

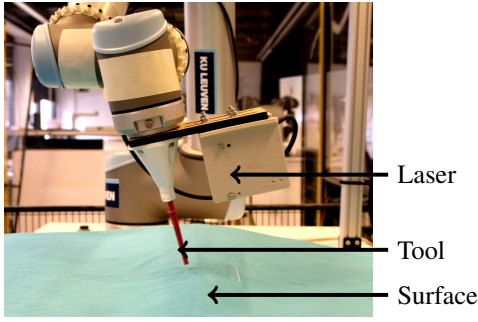
\begin{figure}
    \centering
    \begin{tikzpicture}
        \node[anchor=south west,inner sep=0] (image) at (0,0) {\includegraphics[width=0.27\textwidth]{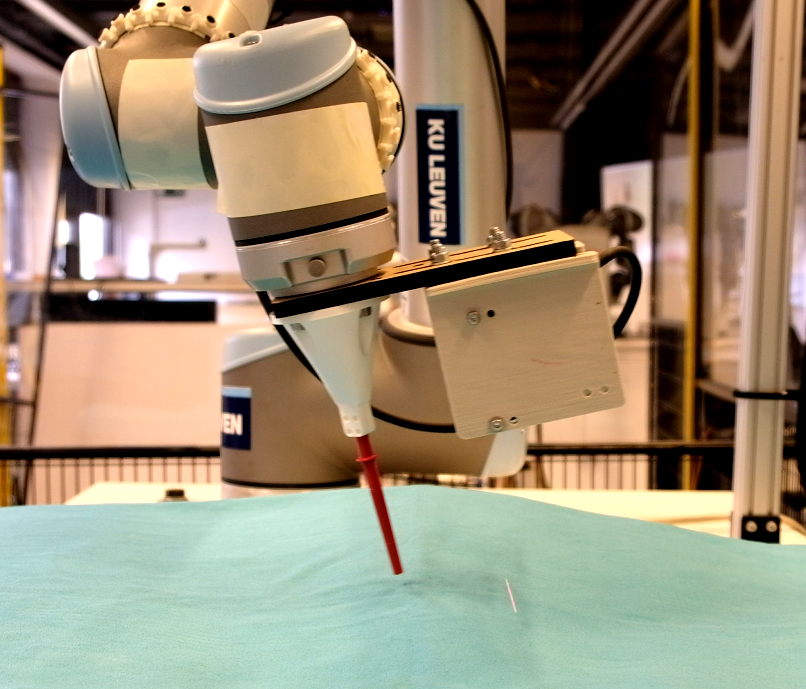}};
        \node[right, yshift=-0.2cm, xshift=0.2cm] (laser) at (image.east) {Laser};
        \draw[->,  line width=1.5pt] (laser.west) -- ++(-2.1,0);
        
        \node[right, yshift=-1.2cm, xshift=0.2cm] (tool) at (image.east) {Tool};
        \draw[->,  line width=1.5pt] (tool.west) -- ++(-2.8,0);

        \node[right, yshift=-1.8cm, xshift=0.2cm] (surface) at (image.east) {Surface};
        \draw[->,  line width=1.5pt] (surface.west) -- ++(-2.4,0);

        \node[left, yshift=0cm, xshift=-1.2cm] (shift) at (image.west) {};

        \node[above, yshift=0em, xshift=0cm] (shift) at (image.north) {};

    \end{tikzpicture}
    \caption{Surface-following task where the robot manipulator is controlled to maintain a desired distance w.r.t. the surface based on laser-sensor measurements.}
    \label{fig:setup_sideways}
\end{figure}

As depicted in Fig.~\ref{fig:setup_sideways}, the goal of surface following is to guide a 6-DOF robot manipulator (Universal Robot UR10) equipped with a tool over a surface while maintaining a desired distance and orientation with respect to the surface. 
The surface's exact shape is unknown beforehand and is estimated online using a laser sensor (Keyence LJ-G080) mounted on the robot's end-effector. 
Consequently, planning a trajectory once and executing it in open-loop is insufficient.
Instead, a feedback control strategy is used to adapt the control input based on the latest surface shape estimates.

\subsection{Surface model and estimator}

This paper considers a local quadratic surface model $\surface_{\surfaceweights}$:
\begin{multline}   
  \underbrace{a_1 + a_2 p_x + a_3 p_y + a_4 p_x p_y + a_5 p_x^2 + a_6 p_y^2}_{\surface_{\surfaceweights}(p_x,p_y)} = p_z,
\end{multline}
where $(p_x, p_y, p_z)$ denotes a point on the surface, represented in the world frame, and $\surface_{\surfaceweights}$ represents the surface model parameterised by  $\surfaceweights = [a_1, \dots , a_6]^\intercal$. 

\begin{figure}
    \centering
    \includegraphics[width = 0.23\textwidth]{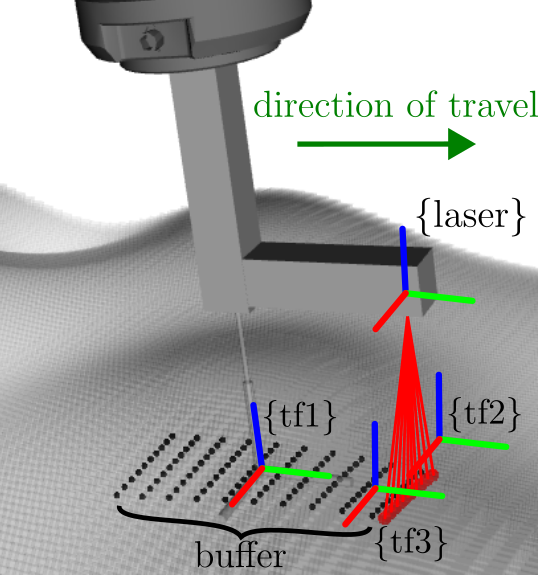}
    \caption{Visualisation of the surface-following scene, showing the laser measurements, buffer of past measurements, and frames introduced for task specification. The axes' \textit{rgb}-colours corresponds with the frames' \textit{xyz}-directions. }
    \label{fig:setup_frames}
\end{figure}

Using the approach in \cite{ubbink2024}, the surface model is estimated online using measurements from the 2D laser sensor.
As shown in Fig.~\ref{fig:setup_frames}, the laser sensor is rigidly attached to the robot's end-effector, and controlled to be in front of the direction of travel.
This placement aims to enhance the accuracy of the estimated surface beneath the tool.
A rolling buffer of $L$ past measurements is maintained that roughly covers the area from the laser sensor to underneath the tool.
At every control interval, the best-fit parameters $\surfaceweights^*$ are estimated from the buffer using the following least-squares approach:
\begin{equation} \label{eq:estimator}
    \surfaceweights^{*} = \arg\min_{\surfaceweights} \sum_{i=1}^{L} [\surface_{\surfaceweights}(p_{x,i}, p_{y,i}) - p_{z,i}]^2.
\end{equation}

\subsection{Task specification} \label{sec:task_spec}

The state, input, and system dynamics are the same as \eqref{eq:acceleration_system}, which assumes an acceleration-controlled robot. 
The task specification was chosen to showcase different ways in which task errors and inequalities constraints can be incorporated into the MPC formulation.
To support the task specification, three additional frames (tf1, tf2, and tf3) are attached to the robot's end-effector, as illustrated in Fig.~\ref{fig:setup_frames}. 
For all tasks a single feedback gain $\alpha = \SI{20}{\per \second}$ is used, whereas different weights are specified ($w_i$ corresponding to diagonal entries in $\mathbf{W}_s$) which determines the relative penalty for deviating from the desired response when conflicting constraints arise. 
The regularisation was selected as $\mathbf{W}_r = 1~\si{\per \radian \squared}$ and $\mu=10^{-6}~\si{\second^{4}}$.
The following task errors and inequality constraints are specified to describe the desired behaviour:
\begin{enumerate}
    \item The position of tf1 ($p_x^\text{tf1}(\state),p_y^\text{tf1}(\state),p_z^\text{tf1}(\state)$) is controlled to remain on the surface:
    \begin{equation} \label{eq:task1}
        e_1(\state) = \surface_{\surfaceweights}(p_x^{\text{tf1}}(\state), p_y^{\text{tf1}}(\state))  - p_z^{\text{tf1}}(\state).
    \end{equation}
    This task error is incorporated in the OCP using the proposed objective function \eqref{eq:pdmpc_objective} with $w_1=10^2~\si{\per \meter \squared}$.
    \item The $x$-position of tf1 is controlled to maintain a fixed reference position ($p_{x,\desired}^\text{tf1} = 0.7~\text{m}$):
    \begin{equation} \label{eq:task2}
        e_2(\state) = 
                p_x^{\text{tf1}}(\state) - p_{x,\desired}^\text{tf1},
    \end{equation}
    and is also incorporated into the OCP with $w_2=10^2~\si{\per \meter \squared}$.
    \item The orientation of the end-effector is controlled such that the laser sensor remains in front of the direction of travel.
    This behaviour is achieved by regulating the $y$-orientation vector of tf1 ($\mathbf{r}_y^\text{tf1}$) such that it remains perpendicular to the world frame's $x$-axis:
    \begin{equation} \label{eq:task4}
        e_3(\state) = \begin{bmatrix}
                    1 & 0 & 0
                \end{bmatrix}
                \mathbf{r}_y^{\text{tf1}} (\state),
    \end{equation}
    with $w_3=10^2~\si{\per \radian \squared}$.
    \item To move the tool over the surface, the desired velocity ($v_{y,\desired}^{\text{tf1}} = 0.15~\text{m\,s}^{-1}$) is controlled along the $y$-axis of tf1:
    \begin{equation} \label{eq:task3}
        e_4(\state) = v^{\text{tf1}}_y (\state)
                -
        v_{y,\desired}^{\text{tf1}}.
    \end{equation}
     This task error is weighted significantly lower than the previous task errors of \eqref{eq:task1}, \eqref{eq:task2}, and \eqref{eq:task4}, with $w_4 = 10^{-2}~\si{\second \squared \per \meter \squared}$.  This choice approximates the behaviour of a hierarchical task-space controller, and prioritises the previous position-level tasks above this velocity-level task when in conflict.
    \item To ensure the laser sensor remains in range, inequality constraints are added such that tf2 and tf3 stay within a desired range ($l_\text{range} = 0.015~\text{m}$) of the surface:
    \begin{subequations} \label{eq:task5}
    \begin{gather} 
        -l_\text{range} \leq \surface_{\surfaceweights}(p_x^{\text{tf2}}(\state), p_y^{\text{tf2}}(\state))  - p_z^{\text{tf2}}(\state) \leq l_\text{range}, \\
        -l_\text{range} \leq \surface_{\surfaceweights}(p_x^{\text{tf3}}(\state), p_y^{\text{tf3}}(\state))  - p_z^{\text{tf3}}(\state) \leq l_\text{range}.
    \end{gather}
    While this is an approximation, it avoids the complexity of computing the actual intersection of the laser beams with the surface.
    These inequalities are incorporated using the proposed constraint formulation \eqref{eq:inequality} with slack variables weighted with $w_5=10^2~\si{\per \meter \squared}$.
    \end{subequations}
    \item Similarly, inequality constraints are enforced on the robot's joint positions and velocities:
    \begin{subequations} \label{eq:task6}
    \begin{gather} 
        \joint^{-} \leq \joint \leq \joint^{+}, \\
        \dot{\joint}^{-} \leq \dot{\joint} \leq \dot{\joint}^{+},
    \end{gather}
    where $\joint^{-}, \joint^{+}$ are based on the specifications of the robot and $[\dot{\joint}^{-}, \dot{\joint}^{+}] = [-1, 1]~\si{\radian \per \second}$. 
    To prevent damage to the robot, these inequalities are enforced as hard constraints (no slack variables).  
    \end{subequations}
    \item Finally, the acceleration of the robot is bounded with:
    \begin{gather} 
        \ddot{\joint}^{-} \leq \ddot{\joint}_\text{cmd} \leq \ddot{\joint}^{+}, \label{eq:joint_accel_lim}
    \end{gather}
    where $[\ddot{\joint}^{-}, \ddot{\joint}^{+}] = [-6, 6]~\si{\radian \per \second \squared} $. 
    Because this constraint contains the control input $\ddot{\joint}_\text{cmd}$, it is directly incorporated into the OCP.
\end{enumerate}
The OCP was modelled using CasADi~\cite{Andersson2019} and solved online using the FAst TRajectory OPtimizer (FATROP)~\cite{vanroye2023fatrop}, an efficient structure-exploiting OCP solver. 
The control and estimation were performed at 100Hz on a desktop PC.

\subsection{Results}

\begin{figure}
    \centering
    \includegraphics[width = 0.47\textwidth]{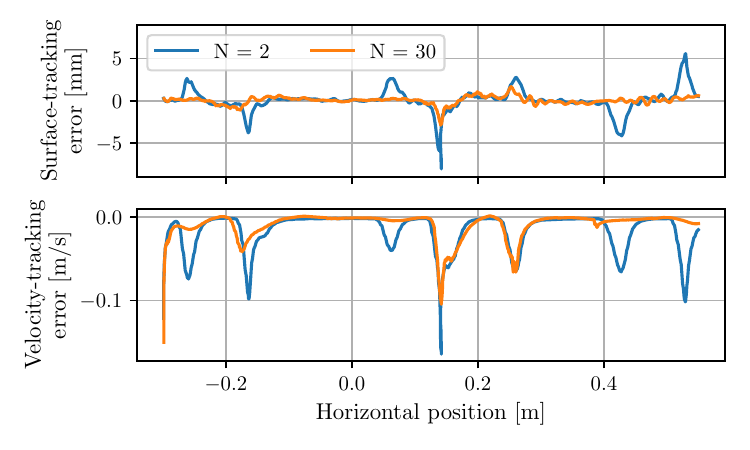}
    \caption{Surface-tracking error (top) and velocity-tracking error (bottom) while moving over the surface.}
    \label{fig:error_vs_progress}
\end{figure}

\begin{figure}
    \centering
    \includegraphics[width = 0.47\textwidth]{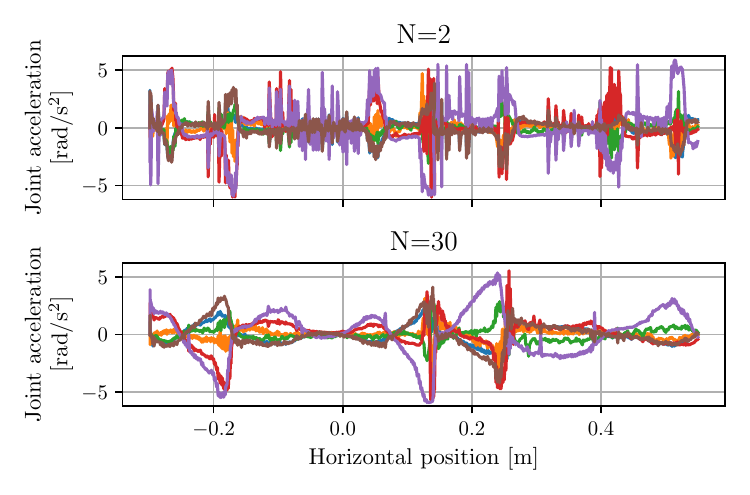}
    \caption{Joint accelerations while moving over the surface.}
    \label{fig:acceleration_vs_progress}
\end{figure}

Fig.~\ref{fig:error_vs_progress} shows the surface-tracking error (top) and the tangential-velocity error (bottom) for horizon lengths $N=2$ and $N=30$ while moving over the surface. 
The corresponding joint accelerations for $N=2$ and $N=30$ are shown in Fig.~\ref{fig:acceleration_vs_progress}.
From these figures, it is clear that the longer prediction horizon not only decreased the tracking error, but also resulted in smoother joint accelerations.

\begin{figure}
    \centering
    \includegraphics[width = 0.47\textwidth]{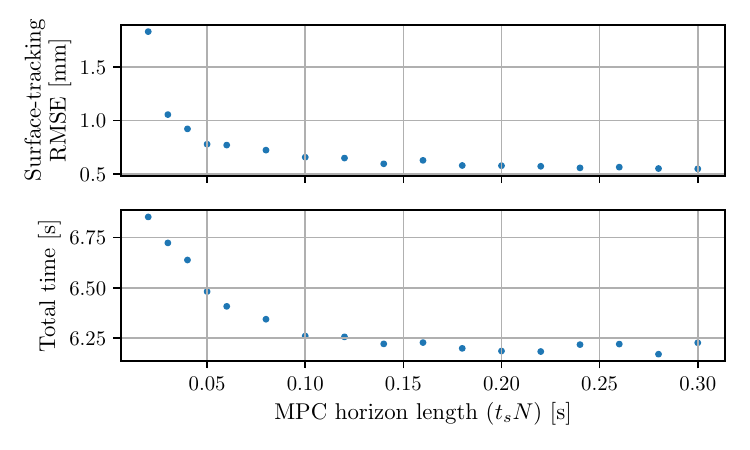}
    \caption{RMSE tracking error and total task time for different horizon lengths.}
    \label{fig:error_vs_horizon}
\end{figure}
\begin{figure}
    \centering
    \includegraphics[width = 0.47\textwidth]{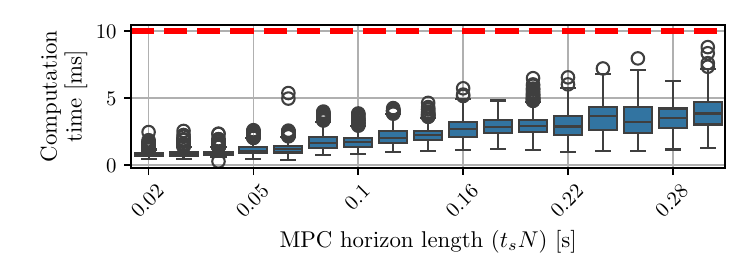}
    \caption{Computation time per horizon length with the dashed red line indicating the maximum allowed time.}
    \label{fig:calctime_vs_horizon}
\end{figure}

Fig.~\ref{fig:error_vs_horizon} summarises the total time taken to move over the surface (bottom) and the surface-tracking root-mean-square error (RMSE) (top) for different horizon lengths.
From the figure, there is a clear trend that the RMSE decreases as the horizon is increased. 
In addition, because the velocity-tracking error improved with the longer horizon, the total time taken to move over the surface decreased by about $10\%$. 
As can be seen in Fig.~\ref{fig:calctime_vs_horizon}, the improved performance did come at the cost of increased computation time. 
However, for all horizon lengths, the computation time remained within the 10 ms sampling time of the control loop.

In summary, the experiments showed that the proposed MPC formulation works for a short horizon of $N=2$, but by extending the prediction horizon: (a) the position tracking error improved, (b) the tool moved faster over the surface, and (c) smoother joint accelerations were achieved. 
These results clearly show that the proposed MPC formulation retains the intuitive tuning parameters of the instantaneous control approach while benefiting from a prediction horizon.

\section{Conclusion} \label{sec:conclusion}
We presented an MPC formulation that aimed to address the question raised in Section~\ref{sec:introduction}:
Yes, there is a way to more easily transition from a working instantaneous controller to an MPC controller.
By using the proposed formulation, it is possible to directly specify the desired time constant of each task error, retaining the intuition of the instantaneous control approach while improving performance through MPC.
We presented the approach from a practical perspective, emphasising usability and ease of tuning and ease of tuning. 
However, there is potential value in revisiting this approach from a more theoretical perspective in future research.



\printbibliography[
heading=bibintoc,
title={References},
]

@book{Rawlings2017,
  title     = {Model Predictive Control: Theory, Computation, and Design},
  author    = {Rawlings, J.B. and Mayne, D.Q. and Diehl, M.},
  year      = {2017},
  publisher = {Nob Hill Publishing}
}

@book{samson,
  title={Robot Control, the Task Function Approach},
  author={Samson, Claude and Le Borgne, Michel and Espiau, Bernard},
  series={Combinatorial Scientific Computing},
  year={1991},
  publisher={Clarendon Press},
}

@article{bouyarmane_2018,
	title = {On {Weight}-{Prioritized} {Multitask} {Control} of {Humanoid} {Robots}},
	journal = {IEEE Transactions on Automatic Control},
	author = {Bouyarmane, Karim and Kheddar, Abderrahmane},
	year = {2018},
}

@article{ubbink2024,
author = {Johan Ubbink and Ruan Viljoen and Erwin Aertbeli\"en and Wilm Decr\'e and Joris De Schutter},
title ={Contactless Surface Following with Acceleration Limits: Enhancing Robot Manipulator Performance through Model Predictive Control},
journal = {European Control Conference},
year = {2024},
}

@article{vanroye2023fatrop,
journal = {IEEE/RSJ International Conference on Intelligent Robots and Systems},
author = {Vanroye, Lander and Sathya, Ajay and De Schutter, Joris and Decré, Wilm},
title = {FATROP: A Fast Constrained Optimal Control Problem Solver for Robot Trajectory Optimization and Control},
year = {2023},
}

@Article{Andersson2019,
  author = {Joel A E Andersson and Joris Gillis and Greg Horn
            and James B Rawlings and Moritz Diehl},
  title = {{CasADi} -- {A} software framework for nonlinear optimization
           and optimal control},
  journal = {Mathematical Programming Computation},
  year = {2019},
  publisher = {Springer},
}

@ARTICLE{gold:2023,
  author={Gold, Tobias and Völz, Andreas and Graichen, Knut},
  journal={IEEE Transactions on Robotics}, 
  title={Model Predictive Interaction Control for Robotic Manipulation Tasks}, 
  year={2023}}

@ARTICLE{Grandia:2019,
  author={Grandia, Ruben and Farshidian, Farbod and Dosovitskiy, Alexey and Ranftl, René and Hutter, Marco},
  journal={IEEE Robotics and Automation Letters}, 
  title={Frequency-Aware Model Predictive Control}, 
  year={2019}}

@INPROCEEDINGS{Mehndiratta:2018,
  author={Mehndiratta, Mohit and Camci, Efe and Kayacan, Erdal},
  booktitle={IEEE/RSJ International Conference on Intelligent Robots and Systems}, 
  title={Automated Tuning of Nonlinear Model Predictive Controller by Reinforcement Learning}, 
  year={2018}}

@inproceedings{jordana2024,
  TITLE = {{Force Feedback Model-Predictive Control via Online Estimation}},
  AUTHOR = {Jordana, Armand and Kleff, S{\'e}bastien and Carpentier, Justin and Mansard, Nicolas and Righetti, Ludovic},
  BOOKTITLE = {{IEEE International Conference on Robotics and Automation}},
  YEAR = {2024},
}

@article{Ziegler1942OptimumSF,
  title={Optimum Settings for Automatic Controllers},
  author={By J. G. Ziegler and Nathaniel B. Nichols},
  journal={Journal of Fluids Engineering},
  year={1942},
}

@INPROCEEDINGS{etasl,
  author={Aertbeliën, Erwin and De Schutter, Joris},
  booktitle={2014 IEEE/RSJ International Conference on Intelligent Robots and Systems}, 
  title={eTaSL/eTC: A constraint-based task specification language and robot controller using expression graphs}, 
  year={2014}}

@INPROCEEDINGS{Edwards:2021,
  author={Edwards, William and Tang, Gao and Mamakoukas, Giorgos and Murphey, Todd and Hauser, Kris},
  booktitle={2021 IEEE International Conference on Robotics and Automation}, 
  title={Automatic Tuning for Data-driven Model Predictive Control}, 
  year={2021}}

@INPROCEEDINGS{Ames:2014,
  author={Ames, Aaron D. and Grizzle, Jessy W. and Tabuada, Paulo},
  booktitle={53rd IEEE Conference on Decision and Control}, 
  title={Control barrier function based quadratic programs with application to adaptive cruise control}, 
  year={2014}}

@inproceedings{Cefalo:2018,
title = {Sensor-Based Task-Constrained Motion Planning using Model Predictive Control},
year = {2018},
booktitle = {12th IFAC Symposium on Robot Control},
author = {Massimo Cefalo and Emanuele Magrini and Giuseppe Oriolo},
}

@ARTICLE{Primbs:2000,
  author={Primbs, J.A. and Nevistic, V. and Doyle, J.C.},
  journal={IEEE Transactions on Automatic Control}, 
  title={A receding horizon generalization of pointwise min-norm controllers}, 
  year={2000}}

@inproceedings{GrandiaCLF:2020,
  title={Nonlinear Model Predictive Control of Robotic Systems with Control Lyapunov Functions},
  author={Ruben Grandia and Andrew J. Taylor and Andrew W. Singletary and Marco Hutter and A. Ames},
  booktitle ={Robotics: Science and Systems},
  year={2020},
}

@inproceedings{allamaa:2024,
      title={Real-time MPC with Control Barrier Functions for Autonomous Driving using Safety Enhanced Collocation}, 
      author={Jean Pierre Allamaa and Panagiotis Patrinos and Toshiyuki Ohtsuka and Tong Duy Son},
      year={2024},
      booktitle={IFAC Conference on Nonlinear Model Predictive Control},
}

@INPROCEEDINGS{Son:2019,
  author={Son, Tong Duy and Nguyen, Quan},
  booktitle={IEEE 58th Conference on Decision and Control}, 
  title={Safety-Critical Control for Non-affine Nonlinear Systems with Application on Autonomous Vehicle}, 
  year={2019}}

@INPROCEEDINGS{Freeman:1995,
  author={Freeman, R.A. and Kokotovic, P.V.},
  booktitle={Proceedings of 1995 American Control Conference}, 
  title={Optimal nonlinear controllers for feedback linearizable systems}, 
  year={1995},
}

@ARTICLE{Cairano:2010,
  author={Di Cairano, Stefano and Bemporad, Alberto},
  journal={IEEE Transactions on Automatic Control}, 
  title={Model Predictive Control Tuning by Controller Matching}, 
  year={2010}}

@INPROCEEDINGS{Bednarczyk:2020,
  author={Bednarczyk, Maciej and Omran, Hassan and Bayle, Bernard},
  booktitle={2020 IEEE International Conference on Robotics and Automation}, 
  title={Model Predictive Impedance Control}, 
  year={2020}
}

\end{document}